\documentclass{article}
\usepackage{ijcai22}

\usepackage{times}
\usepackage{soul}
\usepackage{url}
\usepackage[hidelinks]{hyperref}
\usepackage[utf8]{inputenc}
\usepackage[small]{caption}
\usepackage{graphicx}
\usepackage{amsmath}
\usepackage{amsthm}
\usepackage{booktabs}
\usepackage{algorithm}
\usepackage{algorithmic}
\usepackage{array,multirow}
\usepackage{subfigure}
\usepackage{multicol} 
\usepackage{amsfonts}
\usepackage{stackengine}
\usepackage{xcolor}
\usepackage{mathtools}

\newcommand\ours{{Multiband VAE}}

\def\Eparam{\theta} 
\def\Dparam{\omega}

\def\comment#1{}

\def\eqref#1{(\ref{#1})}
\def\Beq#1\Eeq{\begin{equation}#1\end{equation}}
\def\Beqo#1\Eeqo{\begin{equation*}#1\end{equation*}}
\def\Beqs#1\Eeqs{\begin{align}#1\end{align}}
\def\Beqso#1\Eeqso{\begin{align*}#1\end{align*}}

\title{\ours{}: Latent Space Alignment for Knowledge Consolidation\\ in Continual Learning}

\author{
Kamil Deja$^1$
\and
Paweł Wawrzyński$^1$\and
Wojciech Masarczyk$^1$\and\\
Daniel Marczak$^1$ \And
Tomasz Trzciński$^{1,2,3}$
\affiliations
$^1$Warsaw University of Technology, \\
$^2$Jagiellonian University,\\
$^3$Tooploox \\
\vspace{3px}
\emails kamil.deja.dokt@pw.edu.pl
}

\newcommand\citep[1]{\cite{#1}}
\begin{document}

\maketitle

\begin{abstract}
We propose a new method for unsupervised generative continual learning through realignment of Variational Autoencoder's latent space.
Deep generative models suffer from \emph{catastrophic forgetting} in the same way as other neural structures. Recent \emph{generative continual learning} works approach this problem and try to learn from new data without forgetting previous knowledge. 
However, those methods usually focus on artificial scenarios where examples share almost no similarity between subsequent portions of data --
an assumption not realistic in the real-life applications of continual learning. 
In this work, we identify this limitation and posit the goal of generative continual learning as a knowledge accumulation task. 
We solve it by continuously aligning latent representations of new data
that we call \emph{bands} in additional latent space where 
examples are encoded independently of their source task.
In addition, we introduce a method for controlled forgetting of past data that simplifies this process.
On top of the standard continual learning benchmarks, we propose a novel challenging knowledge consolidation scenario and 
show that the proposed approach outperforms state-of-the-art by up to twofold across all experiments and the additional real-life evaluation.
To our knowledge,~\ours{} is the first method to show forward and backward knowledge transfer in generative continual learning.\footnote{\url{https://github.com/KamilDeja/multiband_vae}}

\end{abstract}

\section{Introduction}

\begin{figure}[bht]
	\centering
	\includegraphics[width=0.9\linewidth]{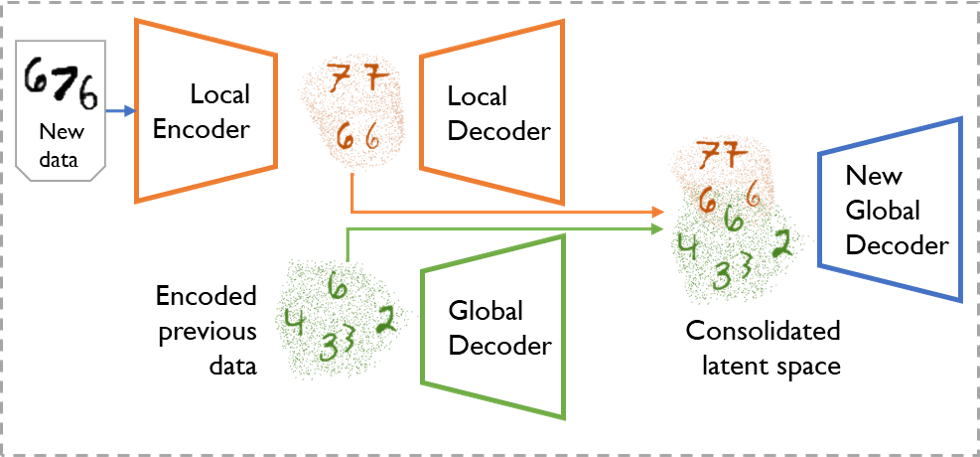}
	\caption{Overview of our \ours{}. With each new task, we first learn a \textit{local} copy of our model to encode new data examples. Then we consolidate those with our current global decoder - main model which is able to generate examples from all tasks.}
	\label{fig:teaser}
\end{figure}

Recent advances in generative models~\citep{2014goodfellow,kingma2014autoencoding} led to their unprecedented proliferation across many real-life applications. This includes
high energy physics experiments at Large Hadron Collider (LHC) at CERN, where they are employed to speed up the process of particles collisions simulations~\citep{paganini2018calogan,deja2020end,kansal2021particle}. 

Those applications are possible, thanks to the main objective of generative methods, which is the modelling of complex data manifolds with simpler distributions.
Unfortunately, this goal remains difficult to deliver in real-life situations where training data is presented to the model in separate portions, e.g., from consecutive periods of data gathering at CERN. The distributions of data within these portions often vary significantly, hence updating model with new examples leads to {\it catastrophic forgetting} of previous knowledge. In generative modelling this is observed through limited distribution of generated examples.  

\emph{Generative continual learning} methods aim to address these challenges 
usually in one of three ways: through regularization~(e.g.~\citep{nguyen2017variational}), adjustment of the structure of a network to the next task ~(e.g.~\citep{rao2019continual}), or rehearsal of previously seen samples when training with new data (e.g.~\citep{rebuffi2017icarl}). 
Nevertheless, practical applications of those methods are yet limited, so most of them focus on the artificial class-incremental (CI) training scenario. 
In this approach, consecutive portions of data (tasks) contain  
disjoint classes and share almost no similarity. While this is the most difficult scenario for discriminative models, we argue that the assumption of classes separation greatly simplifies the problem in generative modelling where task index might be used without reducing the method's generality (detailed discussion in the appendix).

Moreover, the assumption of task independence in CI scenario reduces the complexity of continual learning~\citep{ke2021achieving}. Therefore, in this work, we postulate to investigate the adaptation of generative continual learning methods to the ever-changing data distribution. 
While, for the CI scenario, we expect no forgetting of previous knowledge, in other scenarios, where model is retrained with additional partially similar data, we should aim for performance improvement. 
This can be observed through \emph{forward knowledge transfer} -- higher performance on a new task, thanks to already incorporated knowledge, and  \emph{backward knowledge transfer} -- better generations from previous tasks, when retrained on additional similar examples~\citep{lopez2017gradient}. 

Therefore, to simulate real-life conditions, we prepare a set of diversified continual learning scenarios with data splits following Dirichlet distribution, inspired by a similar approach in federated learning~\citep{hsu2019measuring}. Our experiments indicate that this is indeed a more challenging setup for the majority of recent state-of-the-art continual generative models, which lack sufficient knowledge sharing between tasks.

To mitigate this problem, we propose a \ours{}. The core idea behind our method is to split the process of model retraining into two steps:
(1) a local encoding of data from the new task into a new model's latent space and (2) a global rearrangement and consolidation of new and previous data. 
In particular, we propose to align local data representations from consecutive tasks through the additional neural network. 
In reference to the way how radio spectrum frequencies are allocated, we name data representations from different tasks \emph{bands}. As in telecommunication, our goal is to limit interference between bands. However, we train our model to align parts that represent the same or similar data. 
To support knowledge consolidation between different bands, we additionally propose a controlled forgetting mechanism that enables the substitution of degraded reconstructions of past samples with new data from the current task. 

The main contributions of this work are:
\begin{itemize}
    \item A novel method for generative continual learning of Variational Autoencoder that counteracts catastrophic forgetting while being able to align even partially similar tasks at the same time.
    \item A simple method for controlled forgetting of past examples whenever a new similar data is presented.
    \item A novel knowledge consolidation training scenario that underlines limitations of recent state-of-the-art methods.
\end{itemize}

\section{\label{sec:related}Related Works}
Most of the works incorporating generative models in continual learning relate to generative rehearsal.  
In this technique, the base model is trained with a mixture of new data examples from the current task and recreation of previous samples generated by a generative model. This idea was first introduced by \cite{2017shin+3}, with Generative Adversarial Networks (GAN) trained with the self rehearsal method so-called Generative Replay (GR). 
\cite{lesort2019generative} overview different generative models trained with the GR method. 
Our~\ours{} is a direct extension to this technique.

\paragraph{Continual learning of generative models}
\cite{nguyen2017variational} adapt regularization-based methods such as Elastic Weight Consolidation  (EWC)~\citep{kirkpatrick2017overcoming}, and Synaptic Intelligence (SI)~\citep{zenke2017continual} to the continual learning in generative models regularizing the adjustment of the most significant weights. The authors also introduce Variational Continual Learning (VCL), with adjustments in parts of the model architecture for each task.

In HyperCL, \cite{von2019continual} propose entirely different approach, where a hypernetwork generates the weights of the continually trained model. This yields state-of-the-art results in discriminative models task-incremental training but is also applicable to the generative models. 
In order to differentiate tasks,~\cite{rao2019continual} propose CURL that learns task-specific representation and deals with task ambiguity by performing task inference within the generative model. This approach directly addresses the problem of forgetting by maintaining a buffer for original instances of poorly-approximated samples and expanding the model with a new component whenever the buffer is filled. 
In BooVae,~\cite{egorov2021boovae} propose an approach for continual learning of VAE with an additive aggregated posterior expansion. 
Several works train GANs in the continual learning scenarios either with memory replay~\citep{wu2018memory},
with the extension to VAEGAN in Lifelong-VAEGAN by~\cite{ye2020learning}.

\paragraph{Continual learning with disentanglement}
In VASE by \cite{achille2018life}, authors propose a method for continual learning of shared disentangled data representation.
While encoding images with a standard VAE, VASE also seeks shared generative factors. A similar concept of mixed-type latent space was introduced in LifelongVAE~\citep{ramapuram2020lifelongvae}, where it is composed of discrete and continuous values. In this work we also use a disentanglement method with binary latent space. 
\section{Method}

In this section, we introduce ~\ours{} -- a method for consolidating knowledge in a continually learned generative model. We propose to split generative replay training into two parts: (1) a local training that allows us to build a new data representations band in the latent space of VAE, and (2) global training where we attach a newly trained band to the already trained global model. As a part of the global training, we propose a controlled forgetting mechanism where we replace selected reconstructions from previous tasks with currently available data. 

\subsection{Knowledge Acquisition -- Local Training}

In the local training, we learn a new data representations band by training a VAE using only currently available data. 

Let $\mathbf{x}_{j}^{i}$ denote the $j$-th sample of $i$-th task.
Then, for given sample $\mathbf{x}_{j}^{i}$, and latent variable $\lambda_{j}^{i}$ we use a decoder $p_\Eparam$, which is trained to maximize posterior probability $p(\mathbf{x}_{j}^{i}|\lambda_{j}^{i})$.  
To get the latent variable $\lambda_{j}^{i}$, we use encoder $q_\phi$ parametrized with weights vector $\phi$ that approximates probability $q(\lambda_{j}^{i}|\mathbf{x}_{j}^{i})$. 

To simplify the notation, let us focus on specific task $i$ and drop the index.
As in standard VAE, we follow optimization introduced by~\cite{kingma2014autoencoding} that maximizes the variational lower bound of log likelihood:
\begin{equation}
    \max_{\theta,\phi} \mathbb{E}_{q(\mathbf{\lambda} | \mathbf{x})}[\log p(\mathbf{x} | \mathbf{\lambda})]-D_{K L}(q(\mathbf{\lambda} |\mathbf{x}) \|  \mathcal{N}(\vec{0}, I))).
\end{equation}
where $\theta$ and $\phi$ are weights of encoder and decoder respectively.
In the first task, this is the only part of the training, after which local decoder is remembered as a global one. In other cases we drop local decoder. 
\subsection{Shared Knowledge Consolidation}

\begin{figure}[t!]
\centering
\includegraphics[width=0.75\linewidth]{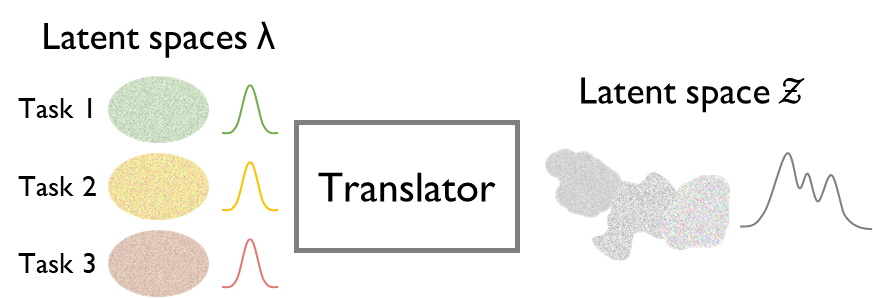}
\caption{Our translator maps individual regularized latent spaces $\lambda$ created by different local models to one global latent space $\mathcal{Z}$, where examples are stored independently of their source task.}
\label{fig:band_alignment} 
\end{figure}

In the second -- global part of the training, we align the newly trained band with already encoded knowledge.
The simplest method to circumvent interference between bands is to partition the latent space of VAE and place new data representation in a separate area of latent space. 
However, such an approach limits information sharing across separate tasks and hinders forward and backward knowledge transfer. 
Therefore, in \ours{} we propose to align different latent spaces through an additional neural network that we call \emph{translator}. Translator maps individual latent spaces which are conditioned with task id into the common global one where examples are stored independently of their source task, as presented in Fig~\ref{fig:band_alignment}. 

To that end, we define a translator network $t_{\rho}(\lambda^{i},i)$ that learns a common alignment of separate latent spaces $\lambda^{i}$ conditioned with task id $i$ to a single latent variable $\mathcal{Z}$
, where all examples are represented independently of their source task. Finally, we propose a global decoder $p_\Dparam(\mathbf{x}|\mathcal{Z})$
that based on distribution approximated with latent variables $\mathcal{Z}$ learns to approximate original data distribution $\mathbf{x}$. 

\begin{figure}[t]
\centering
\includegraphics[width=0.9\linewidth]{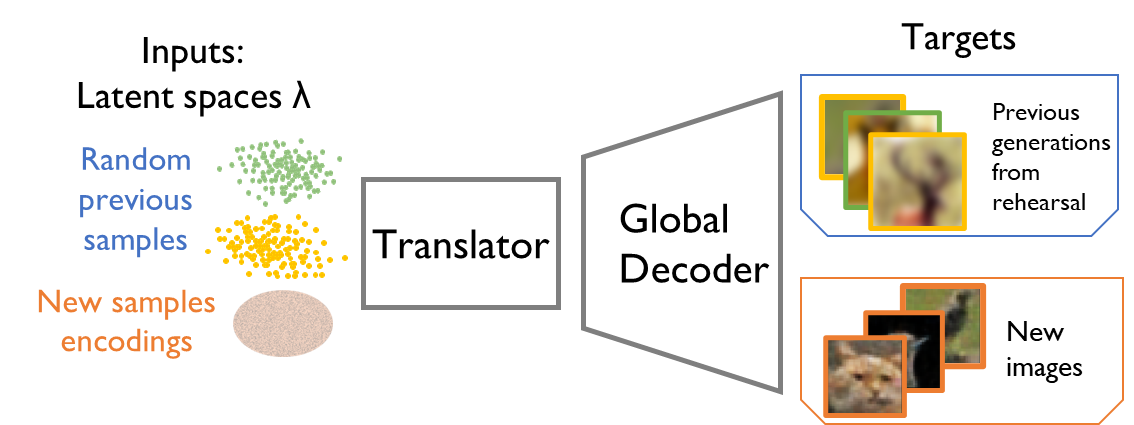}
\caption{We train our translator and global decoder with new data encoded to latent space $\lambda$ associated with original images, and samples of previous data generations generated in a standard rehearsal schema.} 
\label{fig:knowledge} 
\end{figure}

To counteract forgetting, when training translator and global decoder we use auto-rehearsal as in standard generative replay, with a copy of the translator and decoder frozen at the beginning of the task. As training pairs, we use combination of original images $\mathbf{x}$ with their encodings from local encoder $\lambda$, and for previous tasks, random values $\lambda$ with generations $\mathbf{x}$ reconstructed with a frozen translator and global decoder. Fig.~\ref{fig:knowledge} presents the overview of this procedure.

We start translator training with a frozen global decoder, to find the best fitting part of latent space $\mathcal{Z}$ for a new band of data without disturbing previous generations. For that end we minimize the reconstruction loss: 
\begin{equation}
    \min_\rho \sum_{i=1}^{k} ||\mathbf{x}^i- p_\omega(t_\rho(\lambda^i,i))||^2_2, 
\end{equation} 
where $k$ is the number of all tasks.

Then, we optimize parameters of translator and global decoder jointly, minimizing the reconstruction error between outputs from the global decoder and training examples

\begin{equation}
    \min_{\rho,\omega} \sum_{i=1}^{k} ||\mathbf{x}^i- p_\omega(t_\rho(\lambda^i,i))||^2_2, 
\end{equation} 

To generate new example $t$ with~\ours{}, we randomly sample task id $i \sim \mathcal{U}(\{1,\dots,k\})$, where $k$ is the number of all tasks and latent representation $\lambda_{t} \sim \mathcal{N}(\vec{0}, I)$. These values are mapped with translator network to latent variable $\mathbf{z}_{t}$, which is the input to global decoder to generate $\mathbf{x}_{t}$. Therefore, translator and global decoder are the only models that are stored in-between tasks.

\begin{figure}[t!]
\centering
\includegraphics[width=\linewidth]{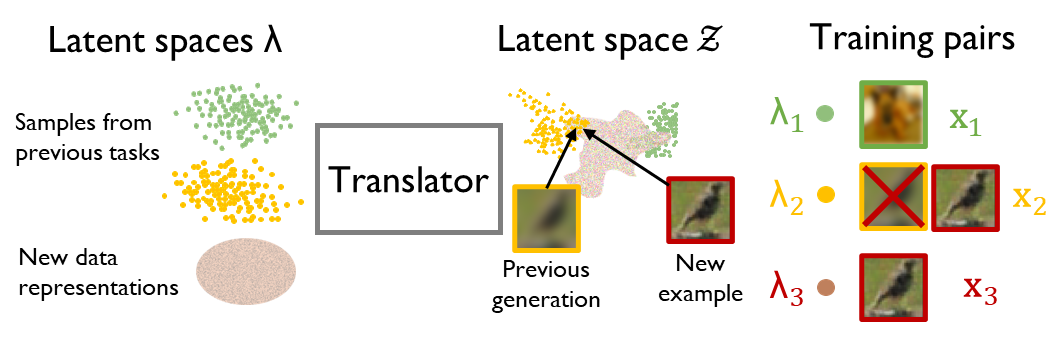}
\caption{When creating rehearsal training pairs with generations from previous data examples, we calculate the similarity between sampled example and the closest currently available data sample in the common latent space $\mathcal{Z}$. If this similarity is above a given threshold, we allow forgetting of the previous reconstruction by substituting the target generation with a currently available similar image.} 
\label{fig:forgetting} 
\end{figure}

\subsection{Controlled Forgetting}

\begin{figure*}[t!]
\centering
\includegraphics[width=0.73\linewidth]{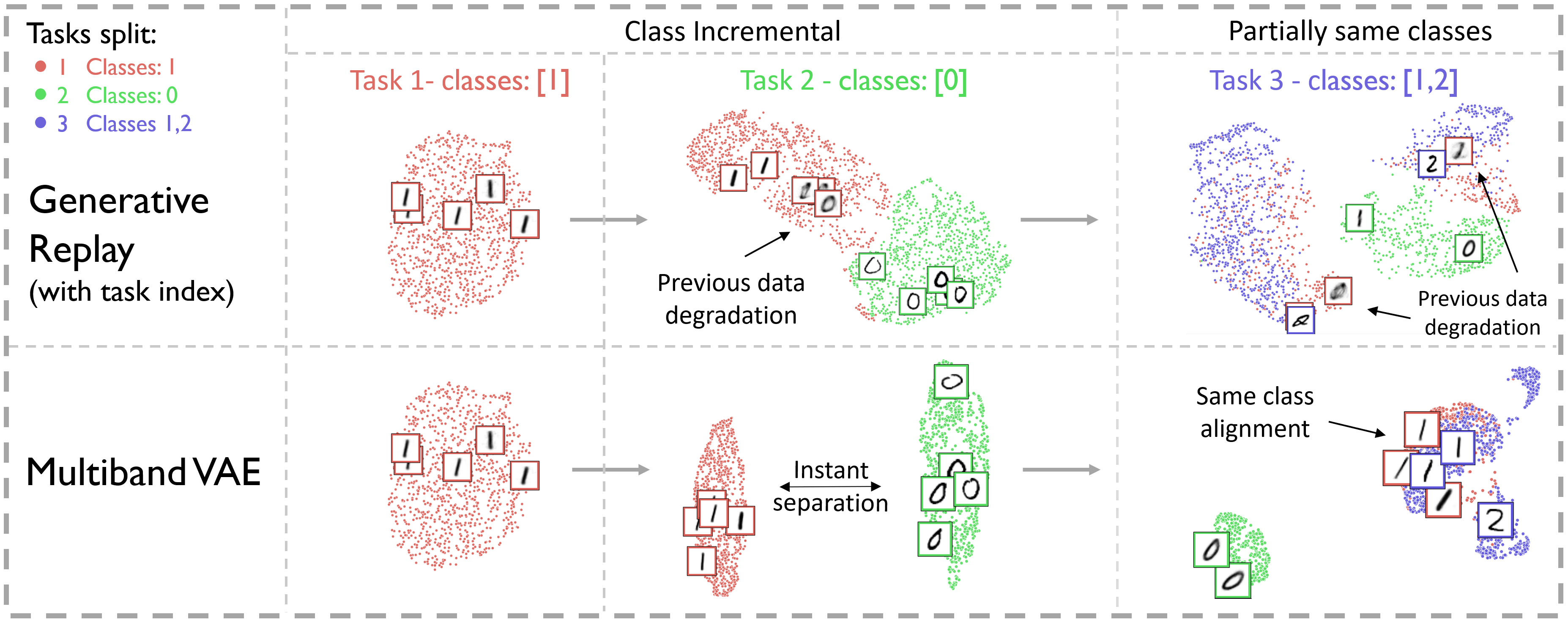}
\caption{Visualization of latent space $\mathcal{Z}$ and generations from VAE in standard Generative Replay and our multiband training for the three tasks (different colors) in a case of entirely different new data distribution, and partially same classes. GR does not instantly separate data from different tasks, which results in the deformation of previously encoded examples. Contrary, our \ours{} can separate representations from different classes while properly aligning examples from the same new class if present.} 
\label{fig:toy_example} 
\end{figure*}

In a real-life scenario, it is common to encounter similar data examples in many tasks. In such a case, we would like our continuously trained model to refresh the memory of examples instead of combining vague, distorted memories with new instances. Therefore, we propose a mechanism for controlled forgetting of past reconstructions during the translator and global decoder joint training. To that end, when creating new training pairs, we compare representations of previous data reconstructions generated as new targets with representations of data samples from the current task in the common latent space $\mathcal{Z}$. If these representations are similar enough, we substitute previous data reconstruction
with the current data sample as presented in Fig.~\ref{fig:forgetting}.

More specifically, when training on task $i$, we first create a subset $\mathcal{Z}^i = t_\rho(q_\phi(\mathbf{x}^i),i)$ with representations of all currently available data in joint latent space $\mathcal{Z}$.
Now, for each data sample $\mathbf{x}^l_j$ generated as a rehearsal target from previous task $l<i$ and random variable $\lambda_j^l$, we compare its latent representation $z_j = t_\rho(\lambda_j^l,j)$ with all elements of set $\mathcal{Z}^i$ 
\begin{equation}
    sim(z_{j}) \coloneqq \max_{z_{q} \in \mathcal{Z}^i} cos(z_{j}, z_{q}).
\end{equation}
If $sim(z_{j}) \geq \gamma$ we substitute target sampled reconstruction $\mathbf{x}^l_j$ with respective original image from $\mathbf{x}^i$.
Intuitively, $\gamma$ controls how much do we want to forget from task to task, with $\gamma=0.9$ being a default value for which we observe a stable performance across all benchmarks.

\section{\label{sec:experiments}Experiments}

To visualize the difference between Generative Replay and \ours{}, in Fig.~\ref{fig:toy_example} we present a toy-example with the MNIST dataset limited to 3 tasks with data examples from 3 classes. 
When presented with data from a new distribution (different class in task 2), our method places a new band of data in a separate part of a common latent space $\mathcal{Z}$. On the other hand, the standard generative replay model learns to transform some of the previous data examples into currently available samples before it can distinguish them, even with additional conditioning on task identity. At the same time, when presented data with partially same classes as in task 3, our translator is able to properly align bands of data representations so that similar data examples (in this case ones) are located in the same area of latent space $\mathcal{Z}$ independently of the source task, without interfering with zeros and twos.

\subsection{Evaluation Setup}
For fair comparison, in all evaluated methods we use a Variational Autoencoder architecture similar to the one introduced by~\cite{nguyen2017variational}, with nine dense layers. However, our~\ours{} is not restricted to any particular architecture, so we also include experiments with a convolutional version. The exact architecture and training hyperparameters are enlisted in the appendix and code repository. We do not condition our generative model with class identity since it greatly simplifies the problem of knowledge consolidation and applies to all evaluated methods. However, similarly to~\cite{ramapuram2020lifelongvae}, we use additional binary latent space trained with Gumbel softmax~\citep{jang2016categorical}. 

\begin{figure}
\centering
\footnotesize
\stackunder[5pt]{\includegraphics[width=0.3\linewidth]{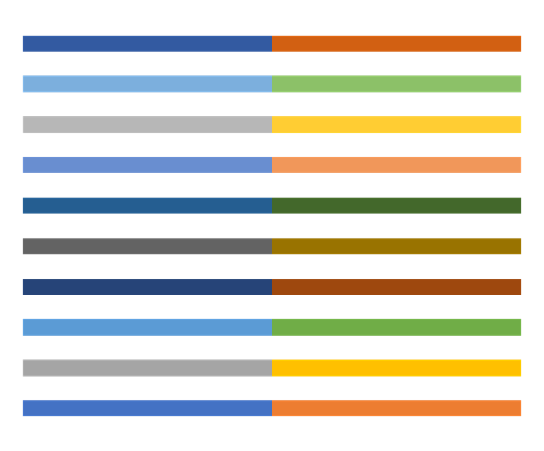}}{Class incremental}%
\hspace{1pt}
\stackunder[5pt]{\includegraphics[width=0.3\linewidth]{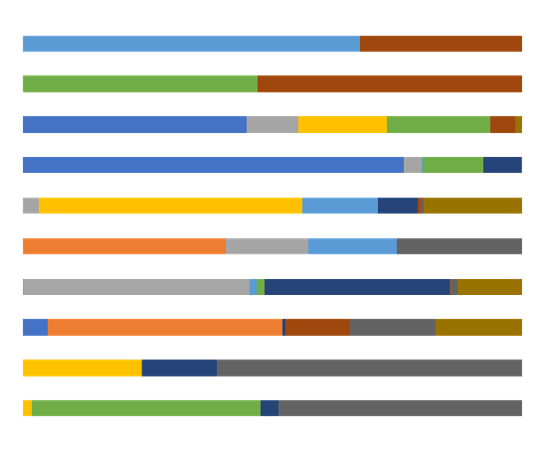}}{Dirichlet $\alpha$=$1$}%
\hspace{1pt}
\stackunder[5pt]{\includegraphics[width=0.3\linewidth]{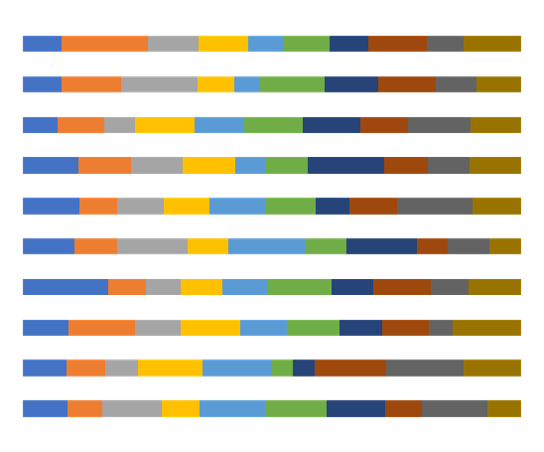}}{Dirichlet $\alpha$=$100$}%

\caption{\small \label{fig:splits} Class splits for different continual learning scenarios. In class incremental split each task consists of separate classes. For $\alpha=1$ Dirichlet distribution, we have highly imbalanced splits with randomly occurring dominance of one or two classes. For higher values of parameter $\alpha$, classes are split almost equally.}
\end{figure}

\begin{table*}[tb]
  \centering
  \resizebox{\textwidth}{!}{
  \begin{tabular}{l||ccc|ccc|ccc|ccc|c}
    \toprule
     & \multicolumn{3}{c}{Split-MNIST}  & \multicolumn{3}{c}{MNIST}& \multicolumn{3}{c}{Split-Fashion MNIST} & \multicolumn{3}{c}{Fashion MNIST} & CERN \\
    & \multicolumn{3}{c}{Class Incremental}&\multicolumn{3}{c}{Dirichlet $\alpha=1$ }&\multicolumn{3}{c}{Class Incremental} & \multicolumn{3}{c}{Dirichlet $\alpha=1$} & Class Inc.\\ 
    Num. tasks &\multicolumn{3}{c}{5}&\multicolumn{3}{c}{10}&\multicolumn{3}{c}{5}&\multicolumn{3}{c}{10}&5\\
    \midrule
    Measure &FID $\downarrow$&Prec $\uparrow$&Rec $\uparrow$ &FID $\downarrow$&Prec $\uparrow$&Rec $\uparrow$ &FID $\downarrow$&Prec $\uparrow$&Rec $\uparrow$ &FID $\downarrow$&Prec $\uparrow$&Rec $\uparrow$ & Wass $\downarrow$\\

    \midrule
    SI & 129 & 77& 80  &153&75&76&134&28&24&140&21&19 &21.1\\
    EWC & 136 & 73& 82 & 120&79&83&126&25&25 &137&24&22& 29.7\\
    Generative replay &120&79&87&254&70&65&96&43&58&133&35&43 & 11.1\\
    VCL & 68 & 85& 94&  127&78& 80&104&30&32 &138&21&20&24.3\\
    HyperCL & 62&91&87 & 148&78&75&108&46&33&155&35&21 &7.8\\
    CURL & 107& 95 &77 & 181&84&74 & 86&47&64 & 83&46&56 & 16.8 \\
    Livelong-VAE & 173 & 75 & 72 &224&63&73 &131&33&62 &201&9&49&7.7\\
    Livelong-VAEGAN& 48 & \textbf{98} & 89 &131 &90 &83 &78 & 54& \textbf{79} &108 &54 &64 &15.1\\
    \textbf{\ours{}}& \textbf{24}&94&\textbf{97}& \textbf{41}&\textbf{92}&\textbf{96}&\textbf{61}&\textbf{66}&69 &\textbf{82}&\textbf{62}&\textbf{65} &\textbf{6.6}\\
    \textbf{\ours{} (conv)}&23& 92&98& 30&92&97 &56&65&72& 77&58&69&8.1\\
    \bottomrule
  \end{tabular}
  }
    \caption{\label{tab:results_mnist} Average FID and distribution Precision (Prec) and Recall (Rec) or Wassserstein distance between original and generated simulation channels, after the final task in different data incremental scenarios. Our method with vanilla architecture outperforms competing solution.
}
\end{table*}

\subsection{Evaluation}
To assess the quality of our method, we conduct a series of experiments on benchmarks commonly used in continual learning (MNIST, Omniglot~\citep{lake2015human}) and generative modeling -- 
FashionMNIST~\citep{xiao2017fashionmnist}. Since the performance of VAE on diverse datasets like CIFAR is limited, in order to evaluate how our method scales to more complex data, we include tests on CelebA~\citep{liu2015faceattributes}. 
For each dataset, we prepare a set of training scenarios designed to evaluate various aspects of continual learning. This is the only time we access data classes, since our solution is fully unsupervised.

\begin{table*}[tb]
  \centering
     \resizebox{\textwidth}{!}{
      \begin{tabular}{l||ccc|ccc|ccc|ccc|ccc}
        \toprule
         & \multicolumn{3}{c}{Split-Omniglot}  & \multicolumn{3}{c}{Split-Omniglot}& \multicolumn{3}{c}{Omniglot} & \multicolumn{3}{c}{FashionM$\rightarrow$MNIST}&\multicolumn{3}{c}{MNIST$\rightarrow$FashionM} \\
        & \multicolumn{3}{c}{Class Incremental}&\multicolumn{3}{c}{Class Incremental}&\multicolumn{3}{c}{Dirichlet $\alpha=1$} & \multicolumn{3}{c}{Class Incremental}&\multicolumn{3}{c}{Class Incremental}\\ 
        Num. tasks &\multicolumn{3}{c}{5}&\multicolumn{3}{c}{20}&\multicolumn{3}{c}{20}&\multicolumn{3}{c}{10}&\multicolumn{3}{c}{10}\\
        \midrule
        Measure &FID$\downarrow$&Prec$\uparrow$&Rec$\uparrow$&FID$\downarrow$&Prec$\uparrow$&Rec$\uparrow$&FID$\downarrow$&Prec$\uparrow$&Rec$\uparrow$&FID$\downarrow$&Prec$\uparrow$&Rec$\uparrow$&FID$\downarrow$&Prec$\uparrow$&Rec$\uparrow$\\
        \midrule
        SI&48&87&81&115&64&28&140&18&16&146&18&15&157&21&19\\
        EWC& 46&88&81&106&68&31 &106&74&38&119&72&30&133&25&23\\
        Generative replay&45&88&82&74&72&62&92&75&53&99&36&45&111&24&39\\
        VCL&48&87&82& 122&62&21&127&71&25&81&45&51&79&45&55\\
        HyperCL& 54&86&76 &98&86&45&115&84&38&128&31&28&143&30&28 \\
        CURL& 22&95&\textbf{95} &\textbf{31}&\textbf{96}&\textbf{92}& \textbf{26}&94&\textbf{92} & 98&\textbf{69}&42 & 122&47&37\\
        Lifelong-VAE& 49&87&83&  79&83&59&  93&83&51& 173&13&50& 200&12&52\\
        Lifelong-VAEGAN& 31&96&90& 71&83&70& 63&85&78& 127&34&61& 91 &52&\textbf{73} \\
        \textbf{\ours{}}& \textbf{21} & \textbf{97} & 93&33& 95&86&41&\textbf{95}&83 &\textbf{51}&65 &\textbf{70}& \textbf{49}&\textbf{67}&\textbf{73} \\
        \textbf{\ours{} (conv)}&12&98&96&24&95&91&24&96&91&49&68&70&49&70&70\\
        \bottomrule
  \end{tabular}
}
    \caption{  \label{tab:results_omni} Average Fréchet Inception Distance (FID) and distribution Precision (Prec) and Recall (Rec) after the final task in different data incremental scenarios. 
    In more challenging datasets \ours{} outperforms competing solutions. 
    }
\end{table*}

To assess whether the model suffers from catastrophic forgetting, we run class incremental scenarios introduced by~\cite{van2019three}. However, CI simplifies the problem of learning data distribution in the generative model's latent space since the identity of the task conditions final generations. Therefore, we also introduce more complex data splits with no assumption of independent task distributions. To that end, we split examples from the same classes into tasks, according to the probability $q\sim Dir(\alpha p)$ sampled from the Dirichlet distribution, where $p$ is a prior class distribution over all classes, and $\alpha$ is a \textit{concentration} parameter that controls similarity of the tasks, as presented in Fig.~\ref{fig:splits}. 
In particular, we exploit the Dirichlet $\alpha=1$ scenario, where the model has to learn the differences between tasks while consolidating representations for already known classes. In such a scenario we expect forward and backward knowledge transfer between tasks.

To measure the quality of generations from different methods, we use the Fréchet Inception Distance (FID)~\citep{heusel2017gans}. 
As proposed by~\cite{binkowski2018demystifying}, for simpler datasets, we calculate FID based on the LeNet classifier pre-trained on the whole target dataset. 
Additionally, we report the precision and recall of the distributions as proposed by~\cite{sajjadi2018assessing}. As authors indicate, those metrics disentangle FID score into two aspects: the quality of generated results (Precision) and their diversity (Recall).

\begin{table*}[t!]
  \centering
  \small
  \begin{tabular}{l||ccc|ccc|ccc|ccc}
    \toprule
    CelebA split& \multicolumn{3}{c}{Class Incremental}&\multicolumn{3}{c}{Dirichlet $\alpha=1$}&\multicolumn{3}{c}{Dirichlet $\alpha=100$}&\multicolumn{3}{c}{Single split}\\ 
    Num. tasks &\multicolumn{3}{c}{5}&\multicolumn{3}{c}{10}&\multicolumn{3}{c}{10}&\multicolumn{3}{c}{1}\\
    \midrule
    Measure &FID$\downarrow$&Prec$\uparrow$&Rec$\uparrow$&FID$\downarrow$&Prec$\uparrow$&Rec$\uparrow$&FID$\downarrow$&Prec$\uparrow$&Rec$\uparrow$&FID$\downarrow$&Prec$\uparrow$&Rec$\uparrow$\\
    \midrule
    Separate models&103&31&21&105&24.5&7.6&109&28.4&10.6&\multirow{3}{*}{88}&\multirow{3}{*}{35}&\multirow{3}{*}{30}\\
    Generative Replay&105&23.4&14.9&109&14.6&7.4&102&17.2&11.6\\
    \textbf{\ours{}}&95&28.5&23.2&93&33&22&89&36.2&28&\\
    \bottomrule
  \end{tabular}
  \caption{\label{tab:results_celeba} Average FID, distribution Precision, and Recall after the final task on the CelebA dataset. Our \ours{} consolidates knowledge from separate tasks even in the class incremental scenario, clearly outperforming other solutions. With more even splits our method converges to the upper bound which is a model trained with full data availability.}
\end{table*}

For each experiment, we report the FID, Precision, and Recall averaged over the final scores for each task separately. For methods that do not condition generations on the task index (CuRL and LifelongVAE), we calculate measures in comparison to the whole test set. The results of our experiments are presented in Tab.~\ref{tab:results_mnist} and Tab.~\ref{tab:results_omni}, where we show scores averaged over three runs with different random seeds. 

To  compare different continual-learning generative methods in a real-life scenario we also use real data from detector responses in the LHC experiment. Calorimeter response simulation is one of the most profound applications of generative models where those techniques are already employed in practice~\citep{paganini2018calogan}. In our studies, we use a dataset of real simulations from Zero Degree Calorimeter in the ALICE experiment at CERN introduced by~\cite{deja2020end}, where a model is to learn outputs of $44 \times 44$ resolution energy depositions in calorimeter. Following~\cite{deja2020end}, instead of using FID, for evaluation, we benefit from the nature of the data and compare the distribution of real and generated channels -- the sum of selected pixels that well describe the physical properties of simulated output. We report the Wasserstein distance between original and generated channels distribution to measure generations' quality. We prepare a continual learning scenario for this dataset by splitting examples according to their input energy, simulating changing conditions in the collider. In practice, such split lead to continuous change in output shapes with partial overlapping between tasks -- similarly to what we can observe with Dirichlet based splits on standard benchmarks (see appendix for more details and visualisations).

\begin{figure}
\centering
\footnotesize
\stackunder[5pt]{\includegraphics[width=0.4\linewidth]{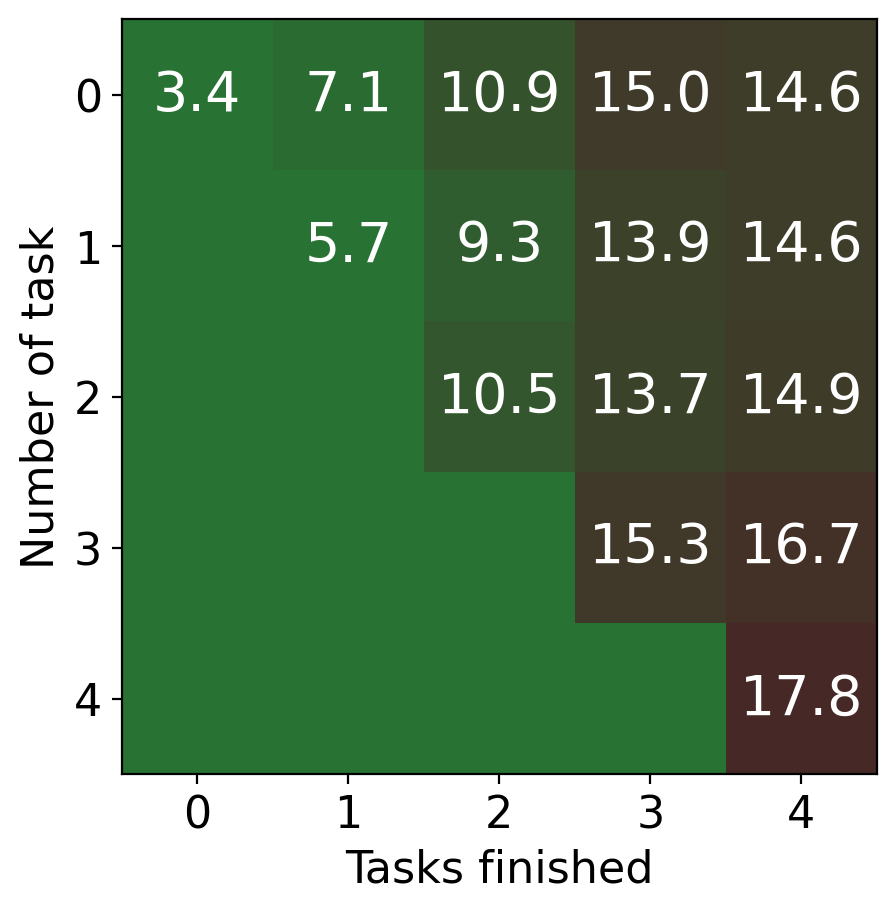}}{\centering Generative replay}%
\hspace{1pt}
\stackunder[5pt]{\includegraphics[width=0.4\linewidth]{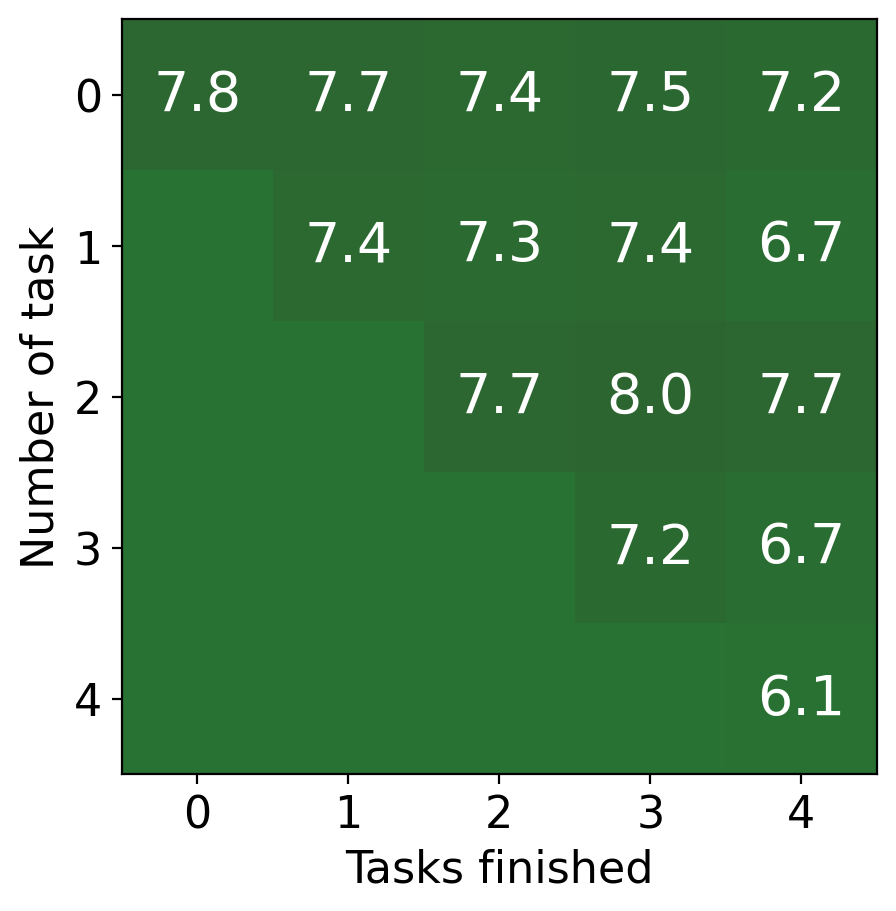}}{\centering \ours{}}%

\caption{\small \label{fig:cern_example} Comparison of Wasserstein distance $\downarrow$ between original simulation channels and generations from VAE trained with standard GR and our multiband training. \ours{} well consolidates knowledge with forward transfer (each row starts with better score) and backward knowledge transfer (improvement for some rows when retrained with more data). At the same time standard GR struggles to retain quality of generations on old tasks.}
\end{figure}

As presented in Tab.~\ref{tab:results_mnist}, our model outperforms comparable methods in terms of quality of generated samples.
Results of comparison on the Omniglot dataset with 20 splits (Tab.~\ref{tab:results_omni}) indicate that for almost all of related methods, training with the data splits according to the Dirichlet $\alpha=1$ distribution poses a greater challenge than the class incremental scenario. However, our~\ours{} can precisely consolidate knowledge from such complex setups, while still preventing forgetting in CI scenario. This is only comparable to CURL that achieves this goal through additional model expansion. 
Experiments on more complex joint datasets, where examples are introduced from one dataset after another, indicate the superiority of~\ours{} over similar approaches. In the real-life CERN scenario, our model also clearly outperforms other solutions. In Fig.~\ref{fig:cern_example} we present how generations quality for this dataset changes in standard generative replay and \ours{}, showing both forward and backward knowledge transfer in \ours{}.

Finally, we evaluate our model with a more complex dataset -- CelebA 
with over 200 000 images of celebrity faces in 64x64 resolution. Based on annotated features, we split the dataset into 10 classes based on the hair color/cover (blonde, black, brown, hat, bald or gray). 
In Tab.~\ref{tab:results_celeba} we show the results of experiments with this dataset split in class incremental and Dirichlet scenarios. 
For class incremental scenario, \ours{} learns to separate bands of examples from different tasks with disjoint distributions, while results improve if in training scenario model is presented with more similar examples. In the latter case, with Dirichlet $\alpha=100$ splits, our model reaches the quality of the upper bound, which is a standard Variational Autoencoder trained with full access to all examples in the stationary training.

\paragraph{Ablation study}
The main contribution of this work is a multiband training procedure, yet we also introduce several mechanisms that improve knowledge consolidation. 
Tab.~\ref{tab:ablation_studies} shows how those components contribute to the final score.

\subsection{Memory Requirements and Complexity}
The memory requirements of \ours{} are constant and equal to the size of the VAE with an additional translator, which is a small neural model with 2 fully connected layers. When training on the new task, our method requires additional temporary memory for the local model freed when finished. This is contrary to similar methods (HyperCL, VCL, CURL) 
which have additional constant or growing memory requirements.
Computational complexity of our method is the same as for methods based on generative rehearsal (VCL, LifelongVAE, Lifelong-VAEGAN). 
In experiments, we use the same number of epochs for all methods, while for~\ours{} we split them between local and global training. 

\begin{table}[tb]
  \centering
  \small
  \begin{tabular}{l|l}
    \toprule
    Modification & FID$\downarrow$\\
    \midrule
    Generative replay & 254 \\ 
    + Two step training &64\\
    + Translator &53\\
    + Binary latent space &44\\
    + Controlled forgetting & 41\\
    + Convolutional model & 30\\
    \bottomrule
  \end{tabular}
\caption{  \label{tab:ablation_studies}
Ablation study on the MNIST dataset with Dirichlet $\alpha=1$ distribution. Average FID  after the last task.}
 
\end{table}

\section{Conclusion}

In this work, we propose a new method for unsupervised continual learning of generative models. We observe that the currently employed class-incremental scenario simplifies the continual learning of generative models. Therefore, we propose a novel, more realistic scenario, with which we experimentally highlight the limitations of state-of-the-art methods. Finally, we introduce a new method for continual learning of generative models based on the constant consolidation of VAE's latent space. To our knowledge, this is the first work that experimentally shows that with continually growing data with even partially similar distribution, we can observe both forward and backward performance improvement. Our experiments on various benchmarks and with real-life data show the superiority of \ours{} over related methods, with upper-bound performance in some training scenarios.

\section*{Acknowledgments}

This research was funded by Foundation for Polish Science (grant no POIR.04.04.00-00-14DE/18-00 carried out within the Team-Net program co-financed by the European Union under the European Regional Development Fund) and National Science Centre, Poland (grant no 2018/31/N/ST6/02374 and 2020/39/B/ST6/01511). 

\bibliographystyle{named}
\bibliography{bibliography}

\appendix
\section*{Appendix}
This is a supplementary material complementing our submission in which we present extended visualizations of the experiments with \ours{}, as well as the implementation details for our models. Finally, we show additional generations sampled from our generative model trained in the continual learning scenarios with the complex datasets such as combined MNIST $\rightarrow$ FashionMNIST and CelebA. Together with this work, we also submit a corresponding codebase, as a part of the supplementary material (public repository link hidden for revision). 
\section{Discussion on the usage of task index in \\generative continual learning}
Access to the task code in continual learning of discriminative models simplifies the problem. It is mostly used when taking crucial decisions such as selecting the relevant part of the model for inference, or the final classification decision. In such cases, a need for task code greatly undermines the universality of a solution.

Contrary to the {\it discriminative} models, in {\it generative} case conditioning generation on task index does not influence or simplify the evaluation setting. The goal of a continually learned generative model is to generate an instance modeled on examples from {\it any} of the previous batches. Hence, to use a continually learned generative model in practice, we can randomly sample a task index (provided that it is lower than the total number of seen tasks) the same way we randomly sample input noise to the decoder or generator. 
In fact, training generative models with task index significantly simplifies a class incremental scenario, in which data distributions from separate tasks -- with different classes are easily distinguishable from each other. In such case task index serves as an additional conditioning input imperceptibly leading to the conditional generative model. This limits the universality of proposed generative continual learning method.

\section{Models architectures}
In this section, we describe in detail the architectures of VAE used in our experiments. The same models and hyperparameters can be found in the codebase.

\subsection{Fully connected Variational Autoencoder}

For comparison with other methods we propose a simple VAE architecture with 9 fully connected layers which we use with simpler datasets: MNIST, Omniglot, FashionMNIST, CERN and combined datasets MNIST $\rightarrow$ FashionMNIST and FashionMNIST $\rightarrow$ MNIST.

In the encoder, we use three fully connected layers transforming input of 784 values through 512, 128 to 64 neurons. Afterward, we map encoded images into continuous and binary latent spaces. For MNIST we use continuous latent space of size 8 and additional binary latent with size 4. For FashionMNIST and Omniglot we extend it to 12 continuous and 4 binary values. 

The translator network takes three separate inputs: continuous encodings, binary encodings, and binary codes representing task number. We first process both binary inputs separately through two fully connected layers of 18 and 12 values for task codes and 8 and 12 neurons for binary encodings. Afterward, we concatenate those three inputs: continuous noise from the encoder and two preprocessed binary encodings into a vector of size 32 for MNIST and 36 for Omniglot and FashionMNIST. We further process these values through two fully connected layers of 192 and 384 neurons which is the size of the second latent space $\mathcal{Z}$

Our decoder consists of 3 fully connected layers with 512, 1024, and final 784 values. In each hidden layer of the model (except for the outputs of the encoder and translator) we use a LeakyRelu activation and sigmoid for the final one.

\begin{figure*}[t!]
\centering
\subfigure[Original data]{\includegraphics[width=.4\linewidth]{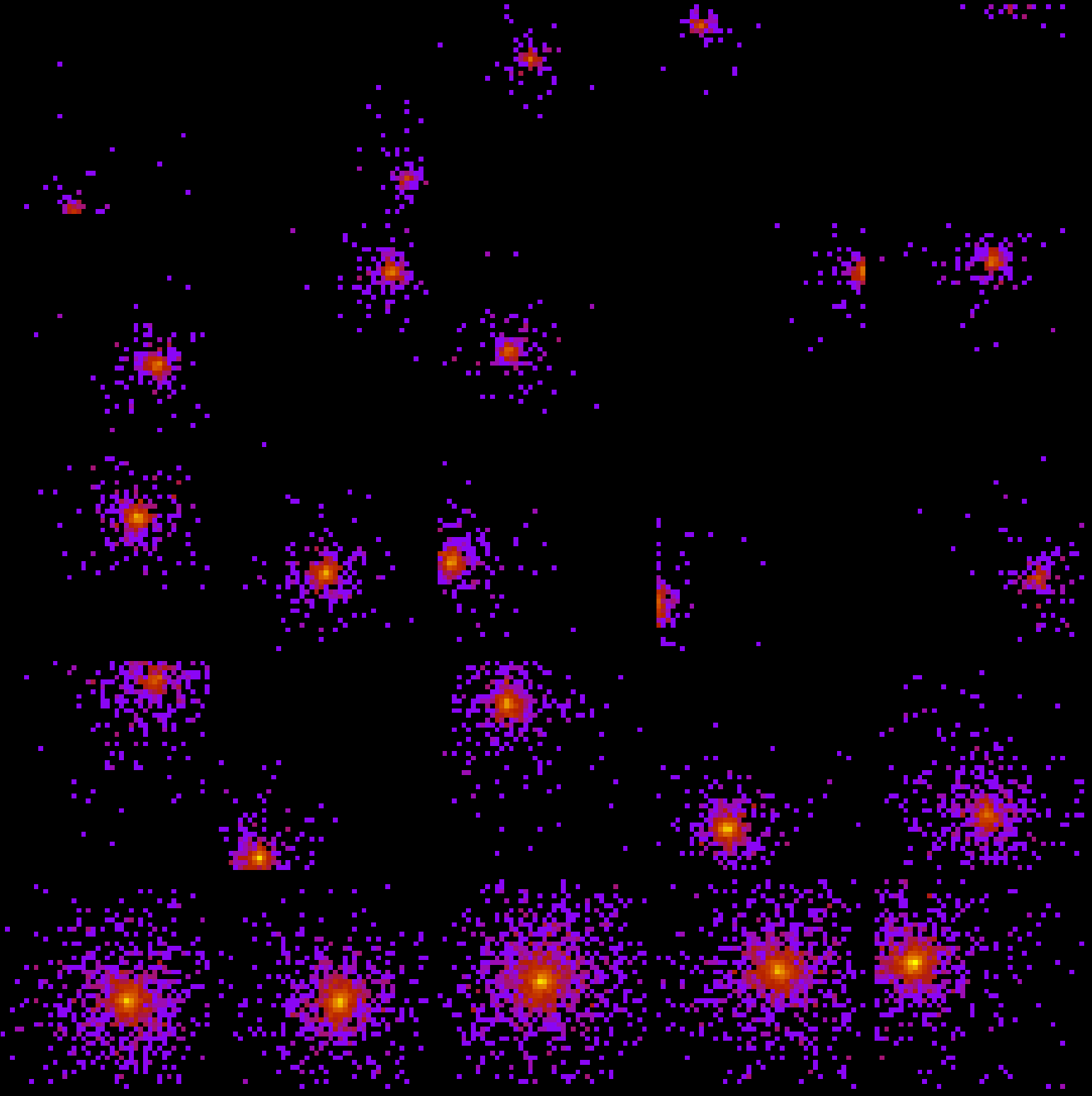}}
\subfigure[Generated simulations]{\includegraphics[width=.4\linewidth]{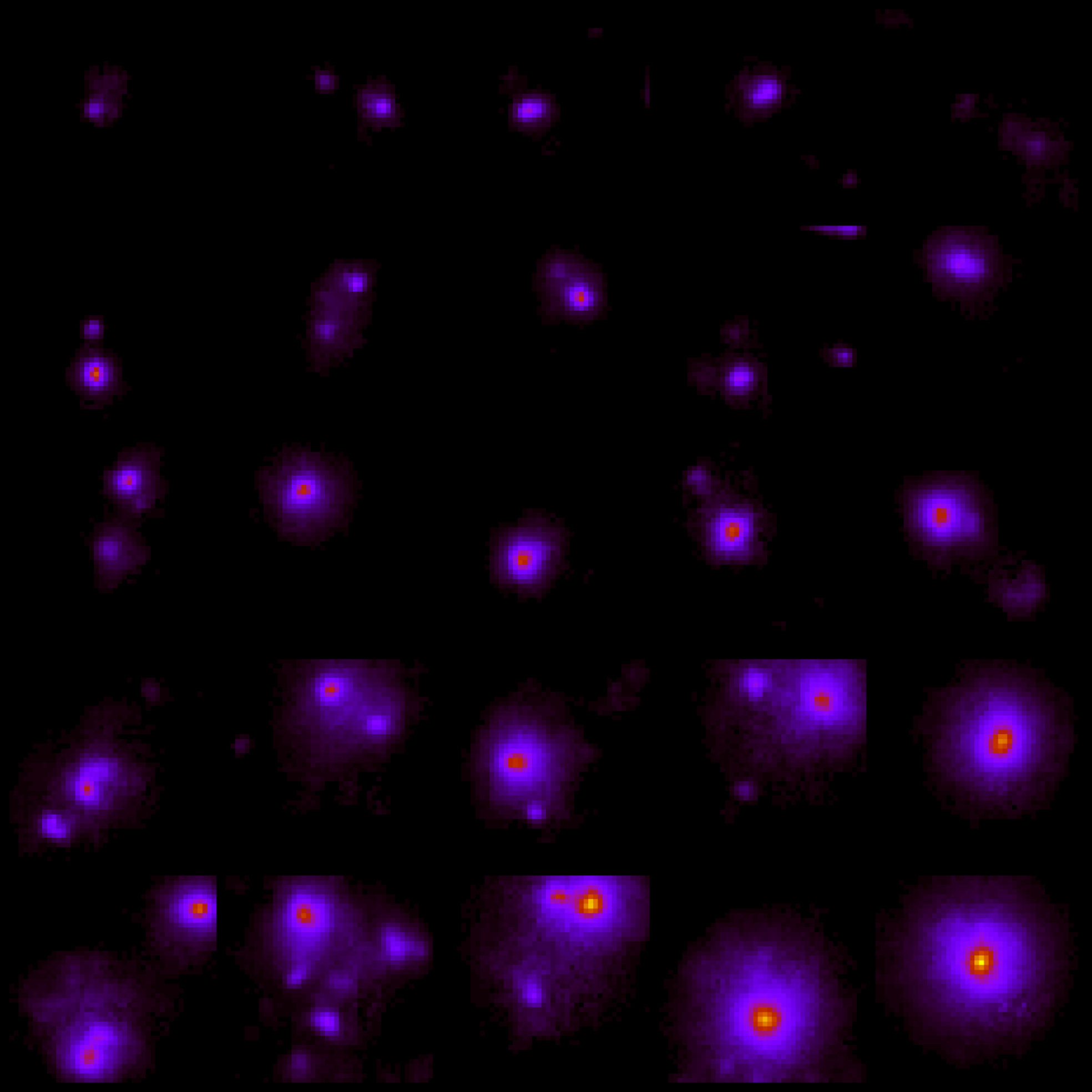}}
\caption{Original simulations for Zero Degree Calorimeter responses and generations from our \ours{} trained in the class incremental scenario on CERN dataset (\textbf{in logarithmic scale}). We split original dataset into 5 tasks (each row of visualisation) with increasing energy of input particle. This results in continuously scaled size of the observed showers with partial overlapping between tasks. \ours{} well consolidates knowledge generating various outputs with full energy spectrum. Although because of the logarithmic scale generated examples seems blurred, this is of the low importance because of the extremely low values of darker/purple pixels.}
\label{fig:cern_input_example} 
\end{figure*}

\subsection{Convolutional VAE}
Our \ours{} is not restricted to any particular architecture, therefore we also include experiments with a convolutional version of our model. In this setup, for the encoder, we use 3 convolutional layers with 32, filers each of $4\times4$ kernel size and $2\times2$ stride. After that, we encode the resulting feature map of 288 features into the latent spaces of the same sizes as in the fully connected model.
For the translator network, we use a similar multilayered perceptron as in the fully connected model, however, we extend the dimensionality of latent space $\mathcal{Z}$ to 512.

In the decoder, we use one fully connected layer that maps the output of the translator with 512 values into 2048 features. Those are propagated through 3 transposed convolution layers with 128, 64 and 32 filters of $4\times4$ kernel size and $2\times2$, $2\times2$, and $1\times1$ stride. The final transposed convolution layer translates filters into the final output with $4\times4$ kernel. 

For the CelebA dataset, we extend our convolutional model. In the encoder, we use four convolutional layers with 50, 100, and 200 filters with $5\times5$ kernel size, followed by fully connected layer mapping 1800 features, through the layer of 200 neurons into the latent space of 32 neurons and binary latent space of 8 neurons. 
In the translator, we extend the fully connected combined layers into 800 and 1600 features which is a dimensionality of latent space $Z$. Our decoder decodes 1600 features from latent space through 3 transposed convolution layers with 400, 200, and 100 filters into the final output with 3 channels.

As in the fully connected model, we use LeakyReLU activations and additional batch normalization after each convolutional layer.

\subsection{Training hyperparameters}
We train our models with the Adam optimizer, learning rate $0.001$ and exponential scheduler with scheduler rate equal to $0.98$. In our experiments, we train our model for 70 epochs of local training and 140 epochs of global training, with 5 epochs of shared knowledge discovery. We combine each mini-batch of original data examples with generations from previous tasks reaching up to mini\_batch\_size$\times$num\_tasks$\times$0.5 samples per mini batch.
 
For the splits according to the Dirichlet distribution we substitute target generations with cosine similarity greater then $0.95$. For class incremental scenario we set this parameter to 1. Nevertheless, our experiments indicate that lowering this value to $0.9$ does not influence model's performance.

\section{Real life CERN dataset}
In this work we evaluate different continual learning generative methods with real-life example of particle collisions simulation dataset. For that end we use data introduced in~\cite{deja2020end} that consists of 117~817 Zero Degree Calorimeter responses to colliding particles, calculated with the full GEANT4~\cite{incerti2018geant4} simulation tool. Each simulation starts with a single particle with a given properties (such as momenta, type or energy) propagated through the detector with simulation tool that calculates interaction of a particle with detector's matter. In case of calorimiters, the final output of those interactions is a total energy deposited in calorimeter's fibres. In case of Zero Degree Calorimeter at ALICE, those fibres are arranged in a grid with $44 \times 44$ size. To simulate continual learning scenario, we divided input data into 5 tasks according to the input particle's energy as presented in Fig~\ref{fig:cern_input_example}. Such split simulates changing conditions inside the LHC, where energy of collided beams changes between different periods of data gathering.

\section{Two latents Variational Autoencoder}
In the global part of our training, we rely on the regularization of VAE's latent space.
In practice, when encoding examples from distinct classes into the same latent space of VAE, we can observe that some latent variables are used to distinguish encoded class, and therefore they do not follow desired continuous distribution as observed by ~\cite{tomczak2018vae} and~\cite{mathieu2019disentangling}. The extended experimental analysis of this phenomenon can be found in the supplementary material.

Therefore, in this work, we propose a simple disentanglement method with an additional binary latent space that addresses this problem, similar to the one introduced in~\cite{ramapuram2020lifelongvae}.
To that end, we train our encoder to encode input data characteristics into a set of continuous variables $\mu_c$ and binary variables $\mu_b$, which are used to sample vectors $\lambda_c$ and $\lambda_b$ that together form $\lambda$ -- the input to the translator model. 
For the continuous variables, we follow the reparametrization trick introduced by~\cite{kingma2014autoencoding}. To sample vector $\lambda_c$, we train our encoder to generate two vectors: means $\mu_m$ and standard deviations $\mu_{\sigma}.$ 
Those vectors are used as parameters of Normal distribution from which we sample  $\lambda_c \sim \mathcal{N}(\mu_m,\text{diag}(\mu^2_\sigma))$. For binary variables, we introduce a similar procedure based on the Gumbel softmax by~\cite{jang2016categorical} approximation of sampling from Bernoulli distribution. 
Therefore, we train our encoder to produce probabilities $\mu_p$ 
with which we sample binary vectors $L_b \sim B(\mu_p)$. 
To allow generations of new data examples, for continuous values, we regularize our encoder to generate vectors $\lambda_c$ from the standard normal distribution $\mathcal{N}(0,I)$ with a Kullback-Leibler divergence. For binary vectors $\lambda_b$, during inference, we approximate probabilities $\mu_p'$ with the average of probabilities $\mu_p$ for all of the examples in the train-set. We calculate $\mu_p'$ during the last epoch of the local training.
Therefore, to generate new data examples we sample random continuous variables $\lambda_c \sim \mathcal{N}(0,I)$ and binary variables $\lambda_b \sim B(\mu_p')$ and propagate them through the  translator and global decoder.

\section{Analysis of binary latent space}
\begin{figure}[tbh]
	\centering
	\includegraphics[width=0.9\linewidth]{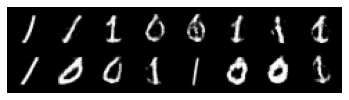}
	\caption{Examples of generations from Variational Autoencoder with no binary latent space, with variables 1 and 2 set to 0. Since model use those variables for class separation, resulting generations with sampled values around $0$ are between two classes.}
	\label{fig:gen_no_bin}
\end{figure}

\begin{figure*}[tbh]
	\centering
	\includegraphics[width=0.9\linewidth]{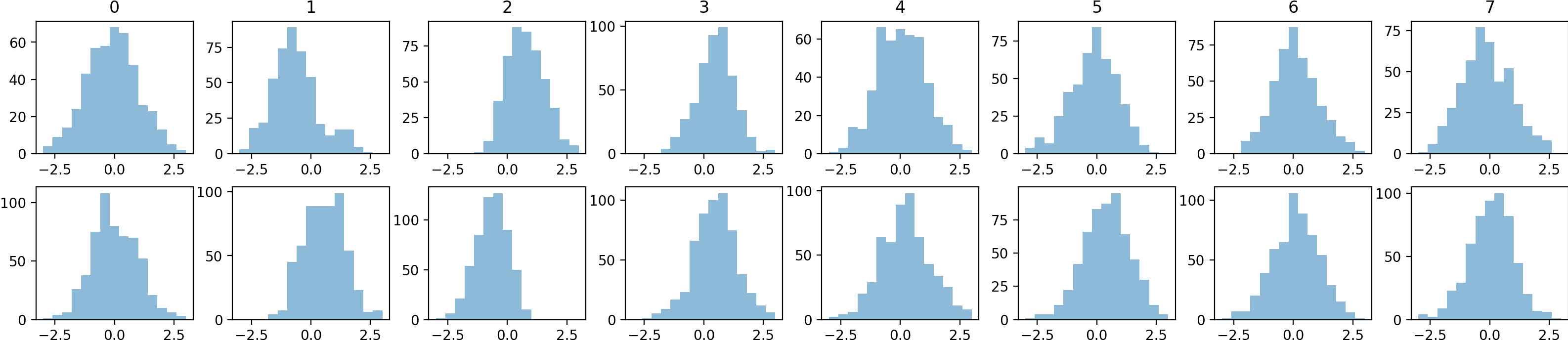}
	\caption{Latent space distribution of Variational Autoencoder trained with two separate classes. Noise embeddings for examples from class zero (top) and one (bottom). Two variables (1 and 2), do not follow standard normal distribution, but are used to differentiate examples from different classes.}
	\label{fig:latent_no_bin}
\end{figure*}

When training Variational Autoencoder with complex data distributions such as a combination of several classes, we can observe that some of the variables in the latent space do not follow desired distribution (e.g. $N(0,1)$), but instead they are used to separate latent space into different parts. In this section, we explain this behavior on the basis of a simple example, with VAE trained on two classes from the MNIST dataset: zeros and ones. For that purpose, we analyze the latent space of the model. In Fig~\ref{fig:latent_no_bin} we present distribution of continuous variables when encoding examples from separate classes. As visible, two variables (1 and 2) do not follow the standard normal distribution to which they were regularized. Instead, they are used to differentiate examples from different classes. Therefore, for certain sampled values, e.g. with variable 2 around 0, the model generates examples that are in between two classes as presented in Fig.~\ref{fig:gen_no_bin}. With generative replay, this problem is even more profound, since rehearsal procedure leads to error accumulation.

\begin{figure}[tbh]
	\centering
	\includegraphics[width=0.99\linewidth]{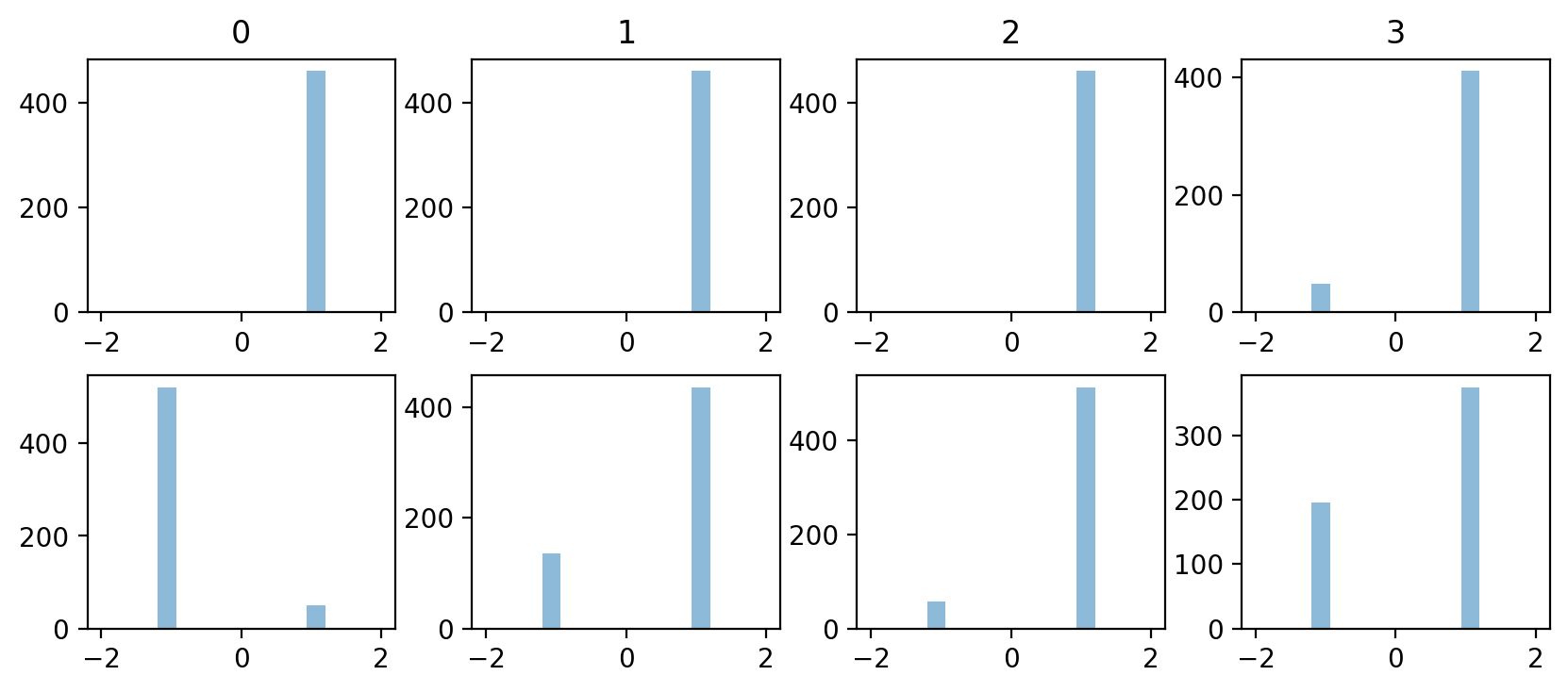}
	\caption{Binary latent space distribution of Variational Autoencoder. Sampled values for examples from encodings of class zero (top) and one (bottom). Additional binary latent space allows for simpler classes separation mostly through the first binary value for which all of the zero examples are encoded with different value than for ones.}
	\label{fig:bin_latent_bin}
\end{figure}

\begin{figure}[tbh]
	\centering
	\includegraphics[width=0.99\linewidth]{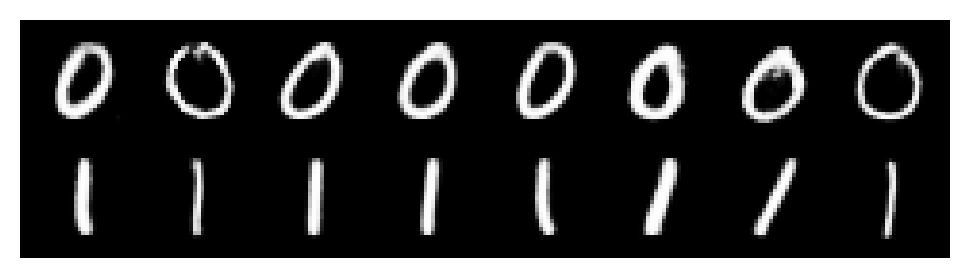}
	\caption{Examples of generations from Variational Autoencoder with binary latent space, for the same random continuous noise (per column) but opposite values for first binary variable. As visible our model well disentangles classes through binary latent space, while continuous values are still used to encode inter-class continuous features such as thickness or rotation.}
	\label{fig:gen_bin}
\end{figure}

\begin{figure*}[tbh]
	\centering
	\includegraphics[width=0.99\linewidth]{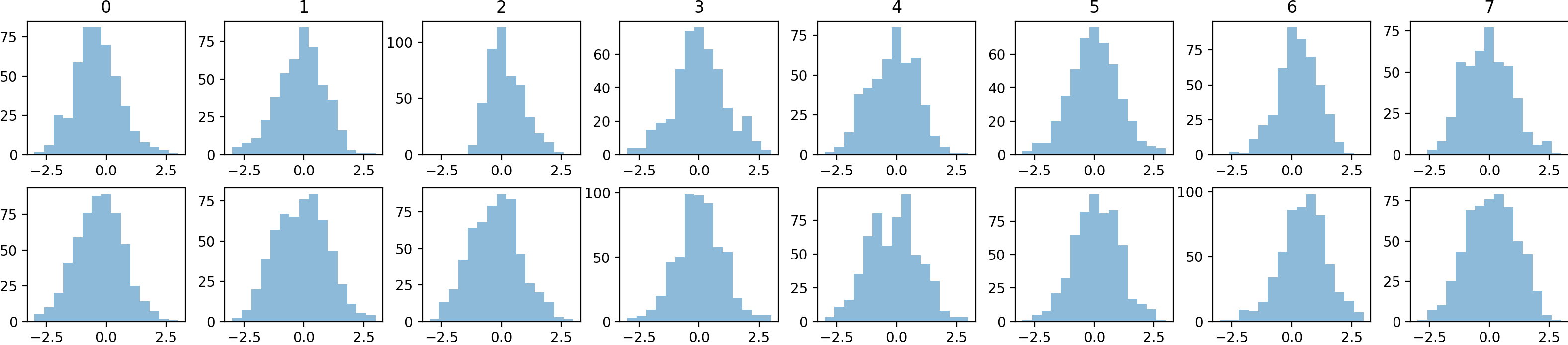}
	\caption{Latent space distribution of Variational Autoencoder with additional binary latent space trained with two separate classes. Noise embeddings for examples from class zero (top) and one (bottom). Thanks to the additional binary latent space, continual variables are better aligned to the standard normal distribution.}
	\label{fig:latent_bin}
\end{figure*}

In this work, we propose a simple disentanglement mechanism. In the process of data encoding, we use an additional binary latent space to which the encoder can map categorical features of the input data such as distinctive classes. This simplifies encoding in standard continuous latent space in which our model does not have to separate examples from different parts of the original distribution. For comparison with standard VAE, we extended the previous model with an additional binary latent space of four binary variables. After training with the same subset of the MNIST dataset of zeros and ones, we observe that model encodes information about classes in the first binary variable as presented in Fig.~\ref{fig:bin_latent_bin}. With such binary codes, our autoencoder does not have to separate classes in the continuous latent space, which leads to better alignment to the normal distribution as presented in Fig.~\ref{fig:latent_bin}. In Fig.~\ref{fig:gen_bin} we show sampled generations from our disentangled representation with two latent spaces. Samples in the same column share the same continuous noise, while those in the same row have the same binary vector.
Visualization indicates that continuous features such as digit's width or rotation are shared between different binary features (column-wise), while for the same binary features (row-wise) we have only examples from the same class. 

\begin{figure*}[t!]
\centering
\subfigure[MNIST~$\rightarrow$~FashionMNIST]{\includegraphics[width=.4\linewidth]{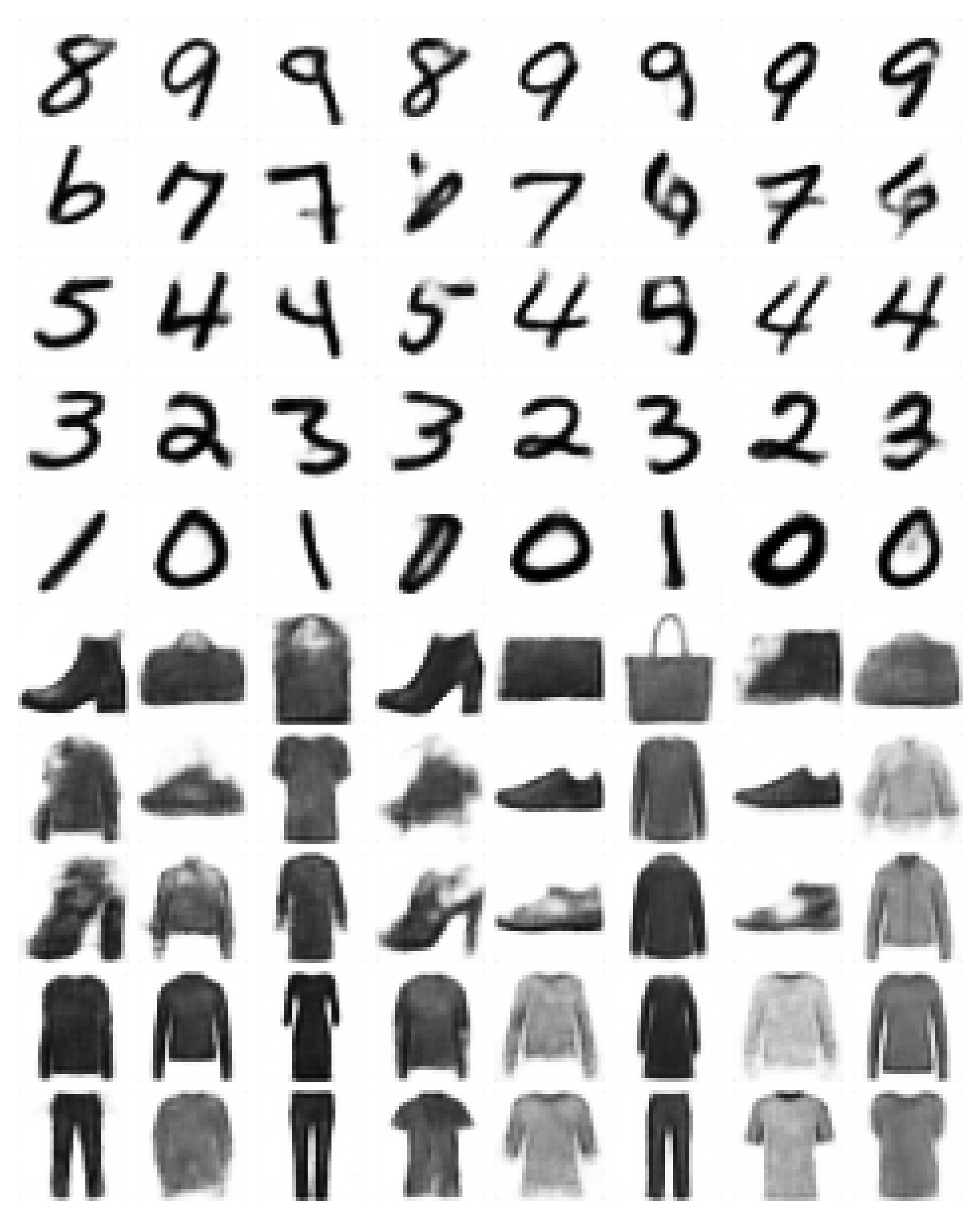}}
\subfigure[FashionMNIST~$\rightarrow$~MNIST]{\includegraphics[width=.4\linewidth]{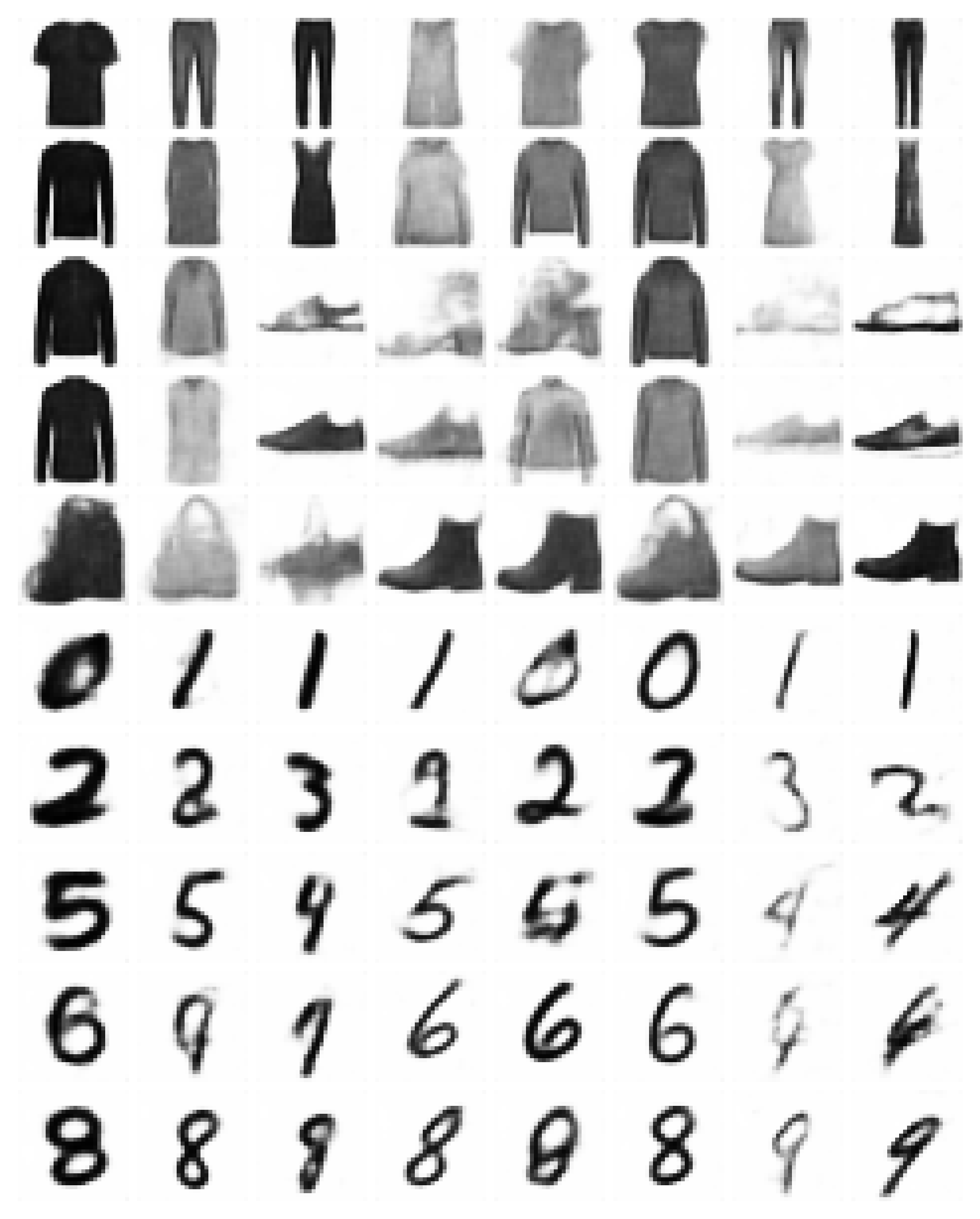}}
\caption{Images generated by our \ours{} trained in the class incremental scenario on combined datasets MNIST~$\rightarrow$~FashionMNIST (left) and FashionMNIST~$\rightarrow$~MNIST (right). We generate images with the same continuous noise per column. Thanks to the proposed band arrangement procedure, we can see that even when trained on drastically different distribution our model adjusts data encodings from various tasks so that they share some common features. For example, in the first column of generations from FashionMNIST~$\rightarrow$~MNIST we can observe how generations of thick black clothes correspond to the firm and bold instances of handwritten digits.}
\label{fig:generations_double} 
\end{figure*}

\begin{figure}[t]
\vspace{3cm}
\centering
\includegraphics[width=0.9\linewidth]{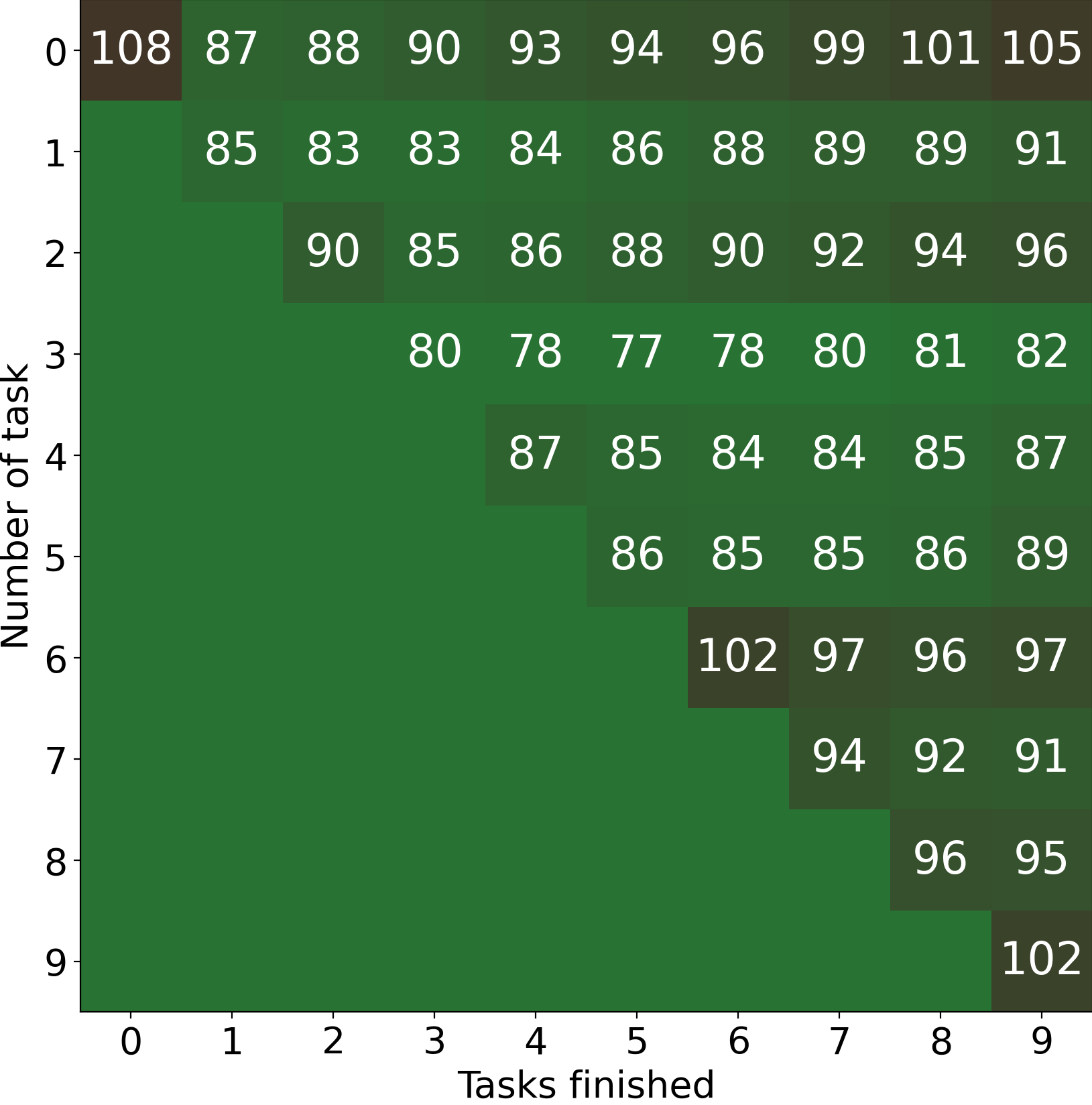}
\caption{ FID$\downarrow$ of generations from a given task of the CelebA dataset, after retraining with number of following tasks for Dirichlet $\alpha=1$ scenario. Our \ours{} well consolidates knowledge with forward and backward knowledge transfer to generations from previous tasks when presented with new similar examples.\vspace{3cm}} 
\label{fig:celeba_example} 
\end{figure} 

\begin{figure}[tbh]
	\centering
	\includegraphics[width=0.7\linewidth]{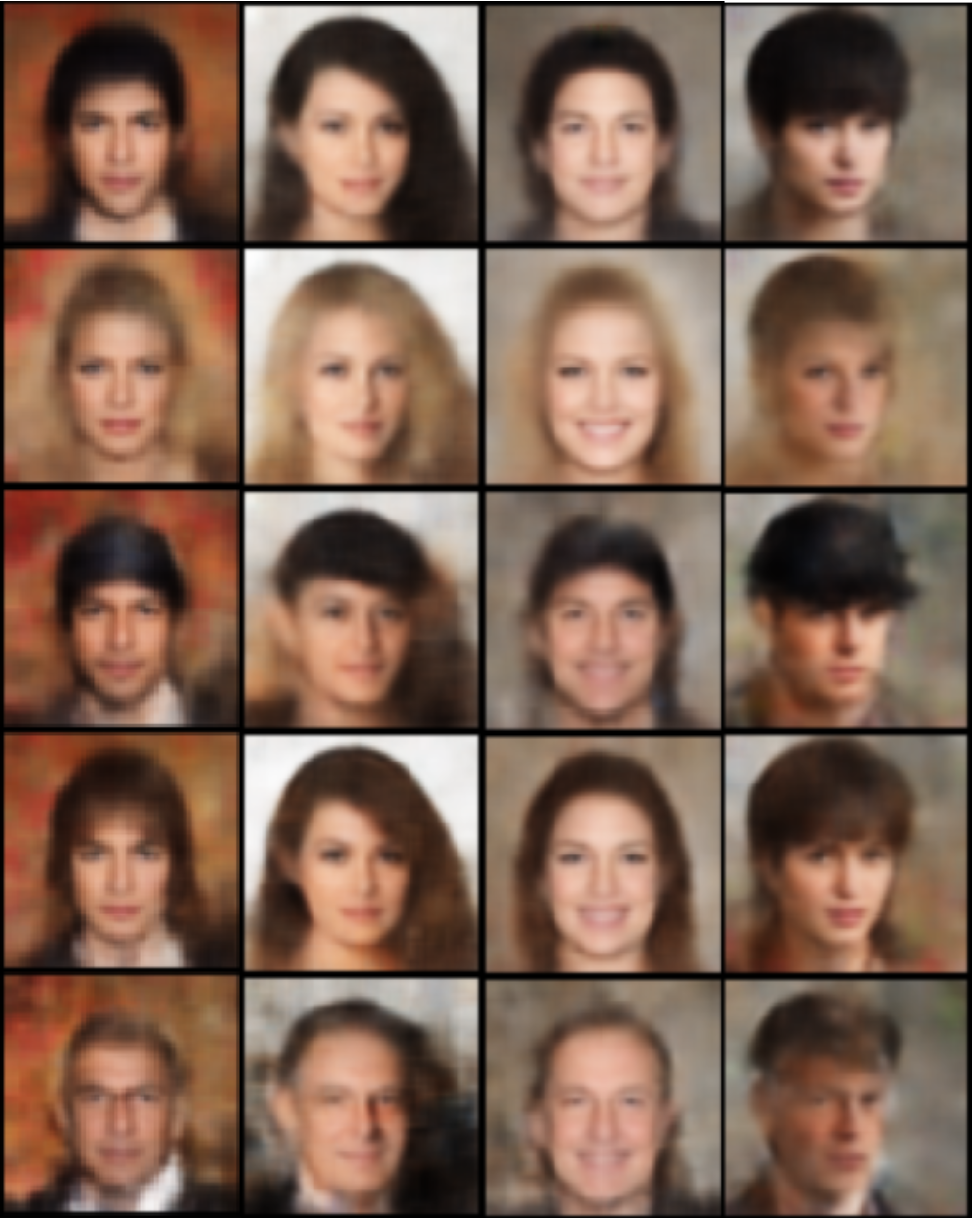}
	\caption{Images generated by \ours{} in the class incremental scenario for CelebA dataset. In the following tasks we introduce images with different hair features. In the first task we introduce photographs of people with black hair, followed by blondes, hats and brown hair. In the final task we train the model with bald and white haired people. In this visualization we present samples with the same random continuous noise (per column) but different task index. We can observe that our \ours{} does not suffer from catastrophic forgetting.}
	\label{fig:geneations_celeba}
\end{figure}

\section{Visualization of generated samples}
In this section, we present additional generations from \ours{}. Fig.~\ref{fig:generations_double} shows generations from combined datasets MNIST~$\rightarrow$~FashionMNIST and FashionMNIST~$\rightarrow$~MNIST. Our model does not suffer from catastrophic forgetting, so previous generations retain their good quality even when retrained with data samples from an entirely different dataset. Moreover, it is able to identify common features between datasets, such as the thickness of generated instances or their general shape. To visualize this behavior we generate samples from the same instance of random continuous noise (column-wise) but conditioned on different task number.

In Fig.~\ref{fig:celeba_example} we present one more example of how our knowledge consolidation works in practice on a standard benchmark. In most cases the quality of new generations from the model retrained on top of the current global models is better than the previous one. Additionally, for some tasks, we can observe backward knowledge transfer in which training on the new task improves generations from the previous ones.

Finally in Fig.~\ref{fig:geneations_celeba} we present generations on the bigger CelebA dataset. Although generations do not match those obtained from state of the art big generative models this is mainly because of the fact that we based our experiments on a shallow model similar to those used in the other approaches (VCL, hypercl, CURL) and other generative autoencoders (WAE, SAE, SWAE). Not to overshadow the main contribution, we did not use additional techniques such as deep models, Laplacian pyramid, or adversarial loss, which would improve the quality of generated samples independently from the training setup.

\end{document}